\documentclass[letterpaper, 10 pt, conference]{ieeeconf}  

\IEEEoverridecommandlockouts                              

\usepackage{amsmath,amssymb,amsfonts}
\usepackage{algorithmic}
\usepackage{graphicx}
\usepackage{textcomp}
\usepackage{xcolor}
\usepackage{url}

\title{\LARGE \bf
Feature-Based Interpretable Reinforcement Learning based on State-Transition Models
}

\author{Omid Davoodi$^{1}$ and Majid Komeili$^{2}$
\thanks{*This work was not supported by any organization}
\thanks{$^{1}$Omid Davoodi - School of Computer Science,
        Carleton University, Ottawa, ON, Canada
        {\tt\small omid.davoudi@carleton.ca}}%
\thanks{$^{2}$Majid Komeili - School of Computer Science,
        Carleton University, Ottawa, ON, Canada
        {\tt\small majid.komeili@carleton.ca}}%
}

\begin{document}

\maketitle
\thispagestyle{empty}
\pagestyle{empty}

\begin{abstract}
\footnote[3]{This work has been submitted to the IEEE for possible publication. Copyright may be transferred without notice, after which this version may no longer be accessible.}
Growing concerns regarding the operational usage of AI models in the real-world has caused a surge of interest in explaining AI models' decisions to humans. Reinforcement Learning is not an exception in this regard. In this work, we propose a method for offering local explanations on risk in reinforcement learning. Our method only requires a log of previous interactions between the agent and the environment to create a state-transition model. It is designed to work on RL environments with either continuous or discrete state and action spaces. After creating the model, actions of any agent can be explained in terms of the features most influential in increasing or decreasing risk or any other desirable objective function in the locality of the agent. Through experiments, we demonstrate the effectiveness of the proposed method in providing such explanations.

\end{abstract}

\section{Introduction}
Recent advances in Artificial Intelligence (AI) have enabled the automation of many real-world tasks. As the current state-of-the-art models are generally based on neural networks, concerns for non-discrimination, transparency, and trust are raised. To address this, Explainable and Interpretable AI methods are used.

One category of Interpretable AI aims to create AI systems that are inherently understandable to human beings. This includes but is not limited to linear models, decision trees, and rule-based systems. One of the drawbacks of these approaches is that there is usually a trade-off between interpretability and model performance. This limits their use in many real-world problems\cite{molnar2020interpretable,adadi2018peeking}.

Another category tries to explain the behavior of an already trained model, treating it as a black box. These so-called post-hoc methods are not without problems but have been shown to offer explanations for complex machine learning models without compromising accuracy or performance.

Reinforcement Learning, one of the areas of machine learning, has also been subject to research for explainability and interpretability. This stems from the fact that AI solutions to many real-world tasks are formulated as RL problems. There are consequences for failure in many of these tasks and as a result, explainability becomes a requirement if such systems are to become operational in the real world. One of the more tangible examples of such tasks would be the autonomous driving of cars.

While there is a sizeable body of research on the explainability of RL, there are fewer works offering a true post-hoc approach for explaining an RL agent. While it is possible to use post-hoc methods developed for explaining general machine learning algorithms like LIME\cite{ribeiro2016should}, the nature of an RL problem could severely limit the usefulness of such methods.

One other factor that is usually forgotten in the domain of the explainability of RL is Risk. Some RL problems are safe in the sense that there is no risk involved in the actions of the agent. On the other hand, some tasks are inherently risky and wrong actions could result in fatal states. Previous research on Safe Reinforcement Learning focuses on creating agents that are robust in terms of risk but does not try to explain the risk itself in a human-understandable manner.

This paper offers a framework and a concrete method for explaining risk in reinforcement learning. The framework requires a preprocessing step that could be done while an agent is being trained in the environment. After this step, the local risk of the actions of any agent in the environment can be explained using the local features of the current state. The method is also independent of the underlying RL method used and requires very little change to the original implementations, as it only requires a log of the state-transitions that happen during the training phase. The proposed method is also not restricted to explaining risk and can be used to offer explanations with respect to any objective function defined over the state space.

It works by creating an estimation of the state transition function of the environment. After this step, the actions of any arbitrary agent in this environment can be explained in terms of important state-space features that are correlated to the overall risk (or the desired objective). Our experiments show that such explanations can be meaningful and understandable.

The main innovation of our work is proposing an abstract framework for human-readable explanations of an objective function in RL in terms of state-space features. We also propose a concrete implementation of this framework and show that this implementation, is able to offer interesting insights about the environment by itself.

\section{Prior Work}

It is theoretically possible to use methods designed for explaining supervised machine learning models to offer explanations for RL models. However, the nature of RL could decrease the value of the results significantly. As a result, several approaches have been proposed for explaining Reinforcement Learning.

One category of such approaches is to solve the problem by developing and using RL methods that are inherently interpretable to human users. This usually translates to using policy functions that are interpretable by themselves. Some approaches tried to implement the policy in some form of decision trees\cite{Bastani2018,rodriguez2019interpretable,roth2019}. Others have used mathematical formulas\cite{Maes2012,hein2018interpretable}, Abstracted Policy Graphs\cite{Topin2019} or fuzzy rule-sets\cite{samsudin2011highly,Hein2017}. The main limitation of such approaches is their dependence on models that often offer inferior performance compared with the state-of-the-art. Sometimes the performance can be increased by increasing the complexity of the models. However, increased complexity usually leads to less interpretability. This issue will be worse in complex problems with high dimensional action and state spaces.

Another approach is to offer explanations using the high-level sub-goals of a Hierarchical Reinforcement Learning method \cite{Beyret2019,Lyu2019}. Hierarchical Reinforcement Learning is a valid tool for increasing the performance of RL agents in an environment with sparse rewards. These methods can offer high-level explanations about the current goals of the agent, but not about the underlying policy itself.

Another direction of research has been to distill the policy function into an interpretable form such as trees\cite{Liu2019} or programming instructions \cite{Verma2018}. This would let the agent keep the original and more performant policy function. The distilled form would then be used to offer explanations to human users. This approach solves the performance-interpretability trade-off but does not address the fact that these explanations would not be very useful when the spaces of the problem are too complex.

Strategy Summarization, first proposed by Amir et al.\cite{Amir2018} tries to approach the problem of explainability from a different angle. It offers explanations in the form of behavior of the agent in interesting states. The rationale behind this is that the user wants to understand how the agent acts in situations where the actions are important, not when the consequences of actions are minimal. Lage et al.\cite{Lage2019} offered a machine teaching approach for extracting interesting states for summarizing the agent's behavior. This approach tries to find a set of state-action pairs that could be sufficient for teaching a new model to behave like an old, already trained one. The main drawback of Strategy Summarization is that it is hard to use in problems that have a large number of interesting states. If the number of explanations is too high, they might overwhelm the human user.

Others have proposed solutions that require certain characteristics from the policy function or the problem itself. These include using reward decomposition for explanations \cite{juozapaitis2019explainable} or using Grad-CAM \cite{selvaraju2017grad} when the policy is in the form of deep neural networks \cite{Weitkamp2019}. This limits their application to general RL problems.

A method for creating local, contrastive explanations was offered by der Waa et al.\cite{VanderWaa2018} in which a state transition model is created as the agent explores the environment. This transition model is then used to predict the outcome of the actions of the current policy compared to a policy given by the human user. The assumption behind this approach is that humans would like explanations only if they have an opinion on how the task should be done. Then implement their opinion in the form of an additional policy and observe the predicted outcome.

One thing common between many of these methods is that they try to explain the individual actions of an agent. In many RL problems, individual actions do not have enough impact on their own. Sometimes, the sequence of actions is more important.

Another drawback is that most of these methods are not scalable. Real-world RL problems can have high-dimensional continuous state spaces. Focusing on single actions will also mean that these approaches will be inadequate in problems with a continuous action space. Our method aims to offer a solution to address these issues.

\section{Proposed Method}
\subsection{Problem Definition}
Consider a \emph{state transition model} which tries to create a model of the transition function $T$ of the RL environment. This model in its purest form is a function $T^*$:$(S\times A\rightarrow S)$ which predicts the next state if the agent takes action $a$ while in state $s$.
Explaining reinforcement learning is a general term that encompasses a range of methods with different goals. In this paper, we consider feature-based interpretability of reinforcement learning where the goal is to explain the risk (or any objective function) involving the agent and its current state.

More formally, given an agent in an environment $E$ with current state $s$, the proposed method aims to find the direction of risk which is a vector $g$ in the state space such that changing state in the direction of $g$ will increase in the agent's objective function $O$:$(S\rightarrow \mathbb{R})$.

This model should be created before explanations are generated. It can be created as the agent is exploring the environment or after the training is done. What is important is that this model will be environment specific and agent agnostic. Once created, the model can be used to explain any agent that operates in this environment.

The other important part of the framework is a definition of the objective function. In this paper, the focus is on risk. To this end, we need a predefined risk in the form of a state-risk function $R^s$:$(S \rightarrow [0,1])$ that determines whether a particular state is risky or not. This function can be determined by a human expert or be learned during the exploration phase. 

Based on the prior work, we can see that there are areas in explainable reinforcement learning that have received less attention. One of these areas is feature-based explanation. For supervised learning, some popular methods like LIME\cite{ribeiro2016should} offer explanations in the form of important features. These types of explanations can be especially helpful when the input space has too many dimensions for the human user to keep track of. Another advantage of LIME is that it can offer local explanations. Local explanations can be easier to understand due to the fact that the model might be very complex globally, but simple in the locality of a single decision.

Explanations are crucial when the decisions of an AI-based system have serious consequences. In RL, consequences are generally in the form of fatal states and risk. Translated to real-world problems, these could be a crash for an autonomous vehicle or fall damage for a robot. A branch of RL called Safe Reinforcement Learning tries to create agents that minimize these risks. But Safe RL does not try to explain the risks of an agent's actions to the human user.

An important aspect that we found during our research was that not all features in the state space are equal in terms of interpretability. We found that for a feature to be a useful explanation, a human should be able to envision action sequences that can cause that feature to change in a relatively predictable manner. In other words, if our explanation is saying that an increase in Feature-A is correlated to increased risk, the human should be able to think of an action sequence that increases or decreases the value of Feature-A for the explanation to be useful for the human user. We call such features \emph{Actionable Features}. If the environment does not have any actionable feature, such features should be constructed out of the state features for our method to be useful.

In essence, we want our method to have these properties:
\begin{itemize}
\item Offer local explanations
\item Offer explanations in terms of the importance of actionable features
\item Offer explanations about the risks the model will be facing
\item Be a post-hoc approach applicable to any arbitrary RL algorithm
\item Require minimal changes to the learning pipeline
\item Be applicable to environments with either continuous or discrete state and action spaces
\end{itemize}

\subsection{Proposed Framework}

\begin{figure}[ht]
\begin{center}
\centerline{\includegraphics[width=\columnwidth]{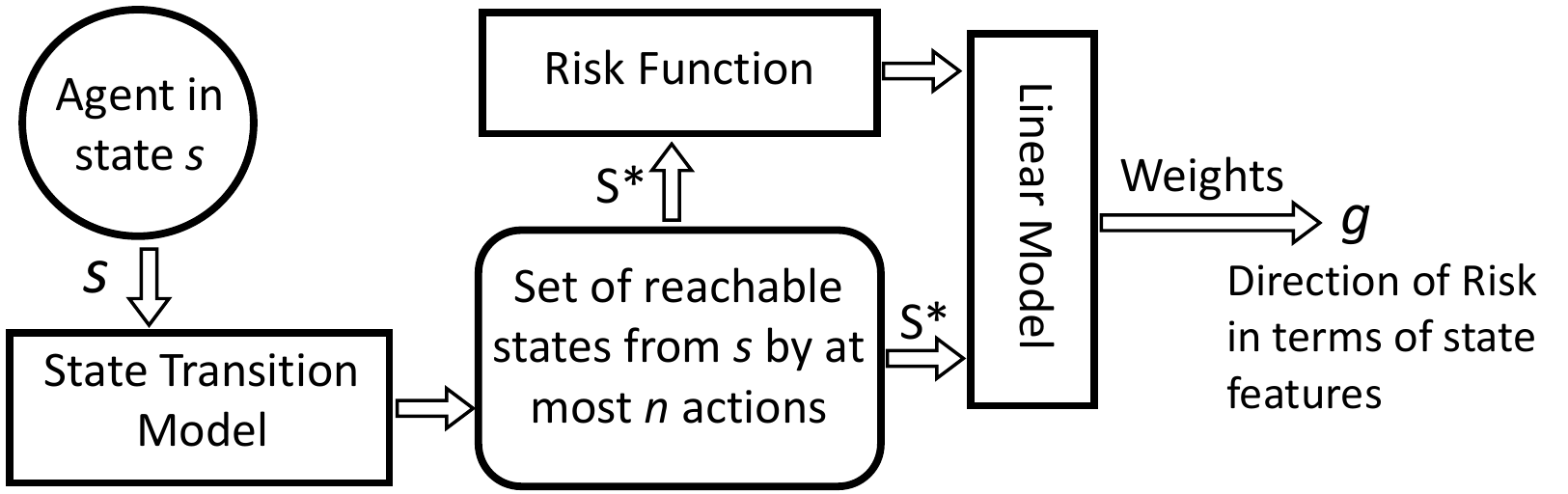}}
\caption{A schema of the proposed framework. For a given state, a set of reachable states is created and used to determine the direction of risk in terms of state features.}
\label{f1}
\end{center}
\end{figure}

To generate the explanations given the current state $s$, we use the state transition model to generate a set of possible states $S^*$ that can be reached by at most $n$ actions when the agent starts from $s$. The objective function $O$ will be defined as the risk for each element of this set which is then estimated using $R^s$. Finally, a linear model is used on the risk estimate. If risk is defined as a binary risky/non-risk concept, a linear classifier is used for classifying the risky states in $S^*$. If risk is a continuous value, linear regression can be used instead. The weights of the linear model can be considered a good candidate for the direction of risk $g$.  Figure~\ref{f1} shows a schema of the proposed framework.

We are assuming that there is a simulation of the environment available where the agent can do exploration in. This is a limiting factor but if there is no such simulated environment available, the method will still be able to function as long as a dataset of state-transitions can be obtained. We do not assume anything about the action or state space. They could be finite, discrete but infinite, bounded or unbounded continuous. Finally, we are assuming that the problem contains some notion of risk. If there is no such notion, our method is not applicable to that problem.

This framework is quite abstract. To test the validity of this framework, we created a concrete implementation. This solution is not necessarily the best possible incarnation of this framework, but gives us a general understanding of this approach and highlights its strengths and weaknesses.

\subsection{Concrete Solution}
In this section, we propose a proof of concept for the proposed framework. The solution and the experiments performed on it are to show the validity of our proposed framework.

For the purpose of this work, by risky states, we mean a binary concept that encompasses fatal and supercritical states as defined by Pecka et al.\cite{Pecka2014}. Fatal states are states where the run (also called an episode) is ended prematurely due to a catastrophic failure. This definition is problem specific. Supercritical states are states that no policy starting from them will prevent the agent from reaching a fatal state. We soften our definition so that supercritical states are states which, starting from these states, no \emph{known} policy exists that will prevent the agent from reaching the fatal states. In this case, we define a state as risky if it is fatal or supercritical.

The next part to determine is the state transition model. For this, we use a state transition graph $G$. This is a directed graph where each node represents a set of states in the state space and each edge from node $n_1$ to node $n_2$ represents the fact that there is at least one action that, starting from one of the $n_1$'s states, leads to one of the $n_2$'s states. One of the states in each node is considered as the representative state of the node. Every state in a node has less distance to the representative state of that node than the representative state of any other node. Note that actions themselves are not explicitly represented in the graph. We admit that this solution is a naive one, but we also think that if our method is able to operate with a crude state-transition graph, then it should also be able to operate with a better constructed one like the ones proposed in \cite{metzen2013learning}\cite{8575599}.

For complex and unbounded environments, we will not be able to create this graph completely. What we propose is to use the interactions of an RL agent with the environment while it is being trained to create this graph in a greedy manner. This agent could be the same RL agent that we are trying to offer explanations about. It is also important to note that the only information we need is the state transitions of the agent while interacting with the environment, not the agent itself. As a result, only a log of the training process is enough for this task if it contains the state transitions. 

For environments with discrete state spaces, this will be trivial. Each node of the graph could represent a single state. As we iterate through the agent-environment interactions, if a state does not have a representative node yet, a new node will be created to represent it. If a transition happened between two nodes that do not have an edge yet, a new edge will be added to the graph.

For environments with a continuous state space, some information will be lost. We define $\epsilon$ as the \emph{radius} of each node. For any state, if there is a node that the new state is within the distance $\epsilon$ of the representative state of that node, then this state is represented by that node. If there is no such node, a new one is created with the current state as the representative state. Edges are created if there is a transition between two states from different nodes.
After constructing the graph $G$, to find the node representing an arbitrary state $s$, we find the node that its representative state has the least distance to $s$.

The risk function is then estimated using $G$. First, any node that represents even one fatal state is set to as risky. Then, any node that all of its outgoing edges lead to risky states are also set as risky. This process continues until no new risky nodes are being set. To determine whether a state is risky or not, we find its representative node in the graph. If that node is risky, then that state is considered to be risky.

To offer explanations, we first find the representative node $n_r$ of the current state $s$. We then find an approximation of $S^*$ by using breadth-first-search (BFS) with a depth limit of $n$ starting from $n_r$. Any node that was traversed by the BFS is then added to a list. This list will be our approximation of $S^*$.

In the end, the linear model is trained on $S^*$. Note that $S^*$ contains nodes and not states. This is to decrease the computational resources needed for training. To train the linear model on the nodes, we used the representative state and risk of that node. After the training of the linear model is finished, the weights of the linear model are considered to be a good solution for $g$. 

We found that the choice of $\epsilon$ is important for the usefulness of the created graph. If it is too small, few states will merge together in graph nodes. In continuous environments where no two states are the same, low $\epsilon$ could cause even close runs to be considered to be completely separate from each other which makes all of the states of unsuccessful runs to be considered supercritical as there is no alternative path detected in the graph. Conversely, all states of successful runs will be considered safe. High $\epsilon$ on the other hand, decreases the expressiveness of the graph by collapsing too many different states into single graph nodes. Good values for $\epsilon$ depend on the problem itself, but we found that the trade-offs were mostly gradual in continuous environments.

The choice of $n$ was also important to the results. If too low, it would not be able to predict much in the future and offer explanations when the agent was very close to risky states. If too high, the explanations would not be local anymore and the direction of risk would be a less reliable indicator. The best value for $n$ depends heavily on how substantial are the effects of single actions on the current state which is problem specific. We found that the results were relatively sensitive to changes in $n$ but the range of meaningful values was small and easy to explore.

\section{Evaluation}

To evaluate our method, we conducted several experiments. In the first experiment, we trained an agent on the Box2D environment bipedal-walker using OpenAI's Gym\cite{brockman2016openai} wrapper. This environment was chosen because it offers an environment where both the state space and the action space are continuous and hence more challenging. It also contains the notion of fatal states in the form of the robot falling on the ground. The RL method used for training the agent is TD3 \cite{fujimoto2018addressing}. All of the interactions of the agent with the environment were logged and used to construct the transition graph with $\epsilon=0.2$ and euclidean distance. For constructing the graph, all of the dimensions of the states were normalized between 0 and 1. The graph had approximately 150,000 nodes, created from approximately 600,000 interactions that were recorded from the TD3 agent used in \cite{fujimoto2018addressing}. Out of those nodes, about 900 were risky based on the definition of risk offered in the first paragraph of III.C.

\begin{figure}[ht]
\begin{center}
\centerline{\includegraphics[width=\columnwidth]{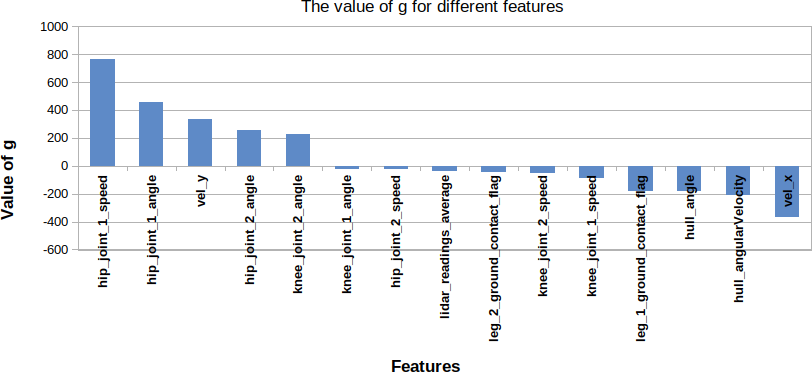}}
\caption{An example of an explanation in the bipedal walker environment.}
\label{f2}
\end{center}
\end{figure}

\begin{figure}[t]
\begin{center}
\centerline{\includegraphics[width=\columnwidth]{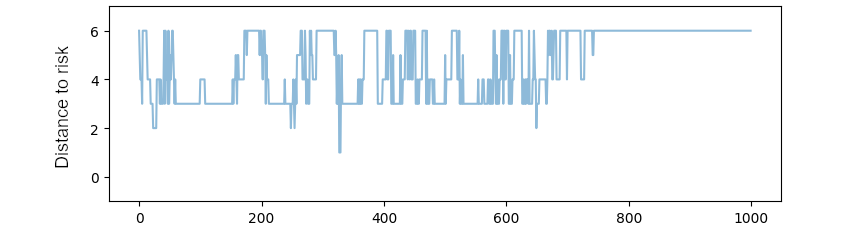}}
\centerline{\includegraphics[width=\columnwidth]{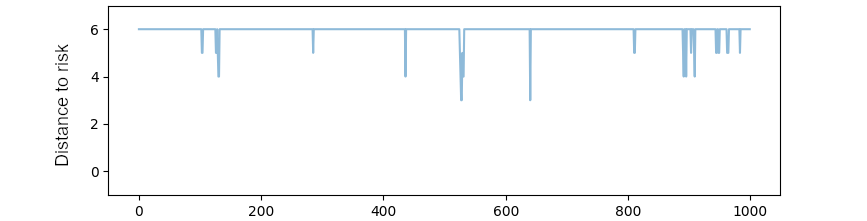}}
\centerline{\includegraphics[width=\columnwidth]{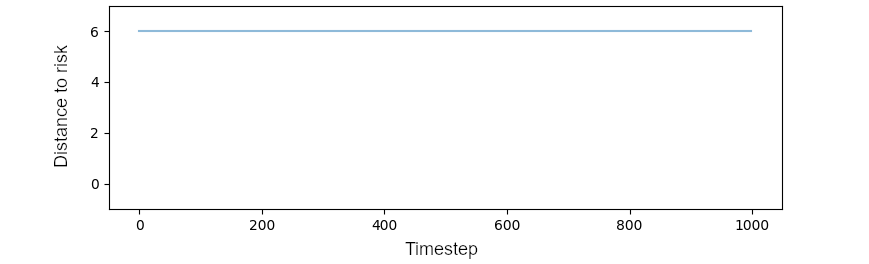}}
\caption{Distance to risky node per timestep during entire non-fatal runs for the three agents in the Bipedal Walker environment. Top: the bad agent, Middle: the better agent, Bottom: the good agent. The vertical axis shows the distance to the nearest risky node and the horizontal axis shows the timestep. Distances greater than 6 are capped at 6.}
\label{f3}
\end{center}
\end{figure}

Figure~\ref{f2} illustrates an example explanation in the Bipedal Walker environment. It can be seen that, in this example, the two most important factors are the angle of the joint between the first leg and the body and its speed. This is because their value for $g$ is higher than the rest of the actionable features. This can be interpreted as \emph{"increasing the angle and speed of the front leg, given the current state the agent is in, will result in a risky situation."}. Given that intuitively, we know that doing so will lead to the robot falling on the ground on its face, the results look promising and could potentially be used for diagnosing RL agents. Also note that some features like hull angle and vertical velocity are hardly actionable and the decrease in horizontal velocity when the robot is falling on the ground is an obvious phenomenon. 

We further demonstrate the application of the proposed method in diagnosing faulty or risky RL agents by offering human interpretable explanations. To this end, the same agent used for the creation of the state-transition graph was put into the simulation environment and non-fatal episodes (runs) were recorded. From these, We create three agents: a "Bad", a "Better", and a "Good" agent. The bad agent was trained only for about 130000 epochs, the better agent was trained for about 320000 epochs, and good agent was trained for about 700000 epochs. 
At each timestep of the run, the minimum distance between the graph node corresponding with the current state of the agent and the nearest risky graph node was calculated. To do so, BFS with a depth limit of 6 was used. If no risky nodes are found within that depth, the current state of the agent is deemed safe for at least 6 actions. This distance could be used as a proxy of how many actions away the agent is from a risky state. Figure~\ref{f3} shows the result of this experiment. The top graph is for the Bad Agent where the vertical axis denotes the distance to the nearest risky node and the horizontal axis is timesteps. For visualization purposes, when no risky node is found in the depth of 6, the distance is still shown as 6. The actual distance could potentially be larger, but computational restrictions hinder our ability to increase the depth much as the process could take exponentially more resources. The middle graph is for the Better Agent and the bottom one shows the results for the Good Agent. As can be seen, there is a noticeable difference between different agents in terms of distance to risk.

\begin{figure}[t]
\begin{center}
\centerline{\includegraphics[width=\columnwidth]{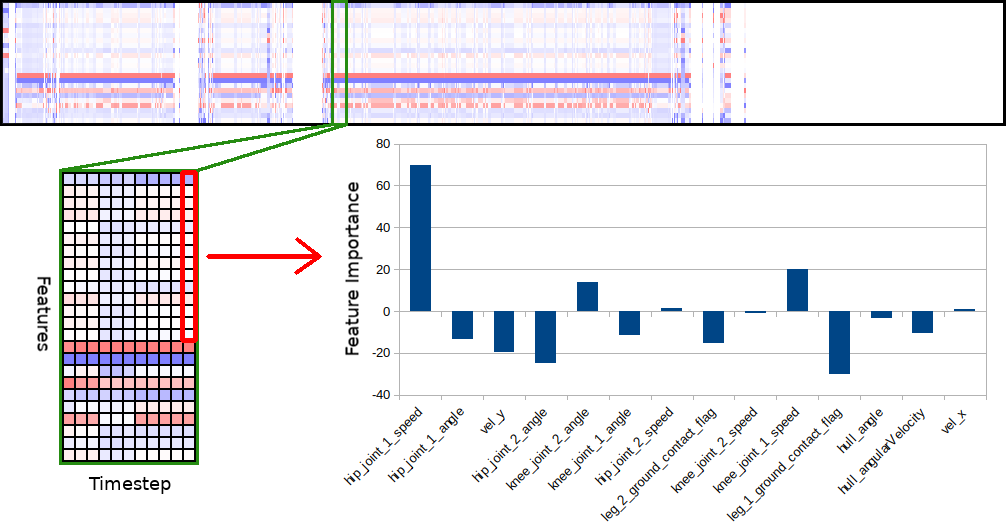}}
\caption{The episode-wide direction of risk graph for a run of the Bad Bipedal Walker agent. Each column represents a state in the episode with colors and their intensity representing $g$, the contribution of each feature to the direction of risk in the environment. The bottom 10 pixels of each column represent the 10 Lidar readings which are not shown here due to space constraints and the irrelevance of the information they provide. Please note that for the ease of visual inspection, the graphic is scaled by a factor of 5 vertically by copying each pixel 5 times. The picture on the left is a zoomed-in version of the area highlighted in green at the top. The chart at the right shows the exact values of $g$ for the column highlighted on the left.}
\label{f6}
\end{center}
\end{figure}

Noticing the usefulness of the episode-wide risk information, we created a graphic that shows the direction of risk $g$ during the entire course of a single run. The graphic is an image where its width is equal to the number of timesteps of the episode in terms of pixels. The height of the image is equal to the number of features representing the state. In this instance, this resulted in a $1000\times24$ pixel image. Each column in this image represents one of the states reached during the course of the run. Each pixel in each column represents the value of each feature in $g$ with more positive values being richer shades of blue and more negative values being richer shades of red. Near-zero values resulted in white or near-white pixels. To get a better visualization, the values of each column were normalized between $(-1, 1)$ such that the column with the most absolute value will be changed to either 1 or -1 depending on it's sign and the rest of the values normalized based on that value. Figure~\ref{f6} shows one of these graphics that was created for the same run that is shown at the top of Figure~\ref{f3}. The figure shows some interesting insights. For example, it shows that while there are variations and exceptions, the overall direction of risk is similar in most of the timesteps. This tells us that the agent, throughout the run, was in an overall position that made it likely to fail in a specific manner. Judging from the actual features themselves especially hip-joint-1-angle and hip-joint-2-angle, this particular manner was the agent falling from behind on the ground. This also shows us that not all features will give us relevant information. The Lidar readings have some of the most intense values in $g$, But these values are not helpful in showing us why an agent fails. The human users are usually able to find the more useful features from $g$ due to their understanding of the environment itself.

\begin{figure}[ht]
\begin{center}
\centerline{\includegraphics[width=\columnwidth]{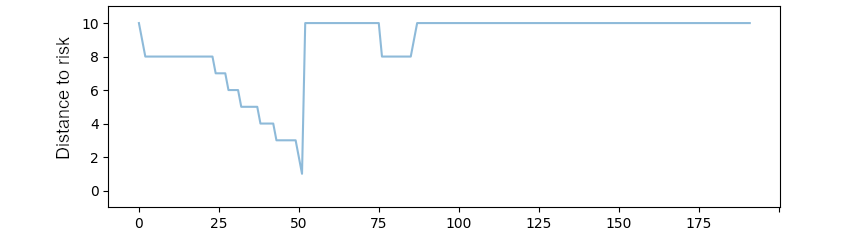}}
\centerline{\includegraphics[width=\columnwidth]{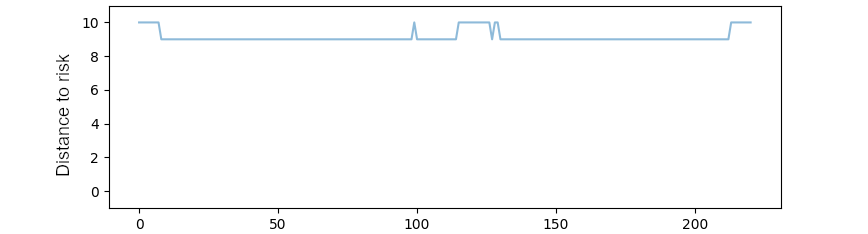}}
\centerline{\includegraphics[width=\columnwidth]{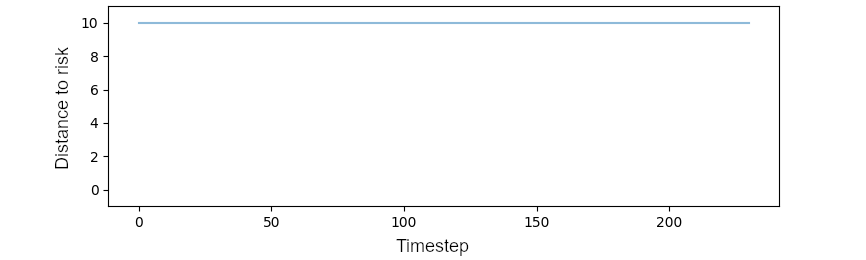}}
\caption{Distance to the nearest risky node during entire non-fatal runs for the three agents in the Lunar Lander environment. Top: the bad agent, Middle: the better agent, Bottom: the good agent. Distances greater than 10 are capped at 10.}
\label{f4}
\end{center}
\end{figure}

A similar set of experiments were performed on the lunar-lander environment from OpenAI Gym. Figure~\ref{f4} shows the results for these experiments. The state-transition graph for this experiment was generated from approximately 400000 interactions with the environment during the training of the same TD3 agent. The resulting graph had $\sim$4000 nodes, $\sim$300 of which were supercritical. An $\epsilon=0.1$ was used. Also $n=10$ was used instead of 6. This is due to the more sparse nature of the state-transition graph that was created for this environment which enabled us to increase the depth while keeping the run time reasonable. The definition of risk was the same as the previous experiment. Please note that fatal states in the Lunar Lander environment are only a subset of failed states. Success in this environment is defined as landing the spacecraft in the designated area without crashing. So, the agent can fail by landing in a wrong spot while not encountering a fatal state. Our base risky states are the ones that only result in a crash (fatal and supercritical states), not the ones that are considered to be mission failure. The reason for this choice was to see how our method would behave in cases where failure and risk are not necessarily the same. The Bad Agent in this scenario was trained for $\sim$10000 epochs. The Better Agent and the Good Agent were trained for $\sim$23000 and $\sim$350000 epochs respectively. The results show that our method is able to offer useful information about the riskiness of an agent even when the risk and failure are not the same concepts.

\begin{figure}[ht]
\begin{center}
\centerline{\includegraphics[width=\columnwidth]{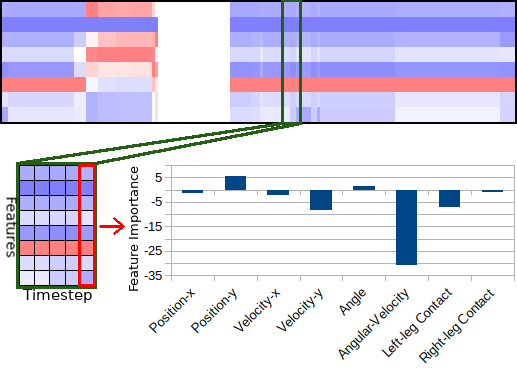}}
\caption{The episode-wide direction of risk graph for a run of the Bad Lunar Lander agent. Each column represents a state in the episode with colors and their intensity representing $g$, the contribution of each feature to the direction of risk in the environment. Please note that for the ease of visual inspection, the graphic is scaled by a factor of 5 vertically by copying each pixel 5 times. The picture on the left is a zoomed-in version of the area highlighted in green at the top. The chart at the right shows the exact values of $g$ for the column highlighted on the left. Dashed area shows a flip in risk direction.}
\label{f7}
\end{center}
\end{figure}

The episode-wide direction of risk graphic was also created for the same run that was recorded from the Bad Agent. The resulting image was $192\times8$ pixels in size. Figure~\ref{f7} shows the resulting image. An interesting point of this image is a near-complete reversal of the direction of risk for a portion of the run (shown in a dashed box). This shows a change of the tilt of the spacecraft which will result in a change in the direction of risk itself (tilting too much left and crashing versus tilting too much right and then crashing). Notice that there are also features such as Position-y that are non-actionable and also irrelevant due to factors like the fact that the site of the crash (the fatal states) will always be on the ground at a narrow range of the vertical position.

\begin{figure}[ht]
\begin{center}
\centerline{\includegraphics[width=100pt]{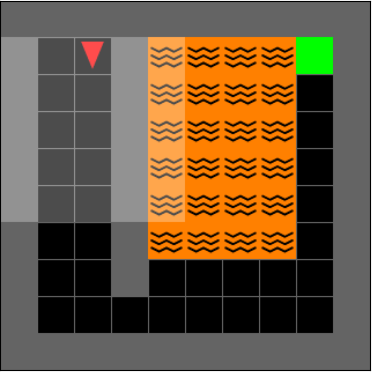}}
\caption{The custom gym-mingrid environment. Black tiles are passable, grey ones are walls, green tile is the goal, orange ones are Lava and the agent is the red triangle currently on the starting position and looking towards to bottom. The observation space is highlighted in lighter colors.}
\label{f8}
\end{center}
\end{figure}

Another test was performed on a modified gym-minigrid environment\cite{gym_minigrid}. There were a number of purposes to this test. To test how the method works in a discrete environment, to test the method in a environment that doesn't have any actionable features, and to validate whether our method offers any advantage over a traditional feature based explanation method like LIME. The environment consists of an agent in a tiled map where some tiles are passable, some are impassable and some are lava (fatal). We created a custom map which is shown in Figure~\ref{f8}. The state space in this environment is a $5\times5$ square of tiles that start from the agent itself and extend 2 tiles to the sides and 4 tiles towards the direction the agent is facing. This means that individual features are the type of terrain each tile contains. As explanations like "making this tile more lava and less wall" are very hard to interpret, we need to create actionable features out of this observation space. We do this by finding the x and y coordinates of the agent relative to it's starting point and direction. The agent can perform 3 actions. One is to go one tile forward and the other two are to rotate 90 degrees to the right or the left. We assume that the initial position is (0, 0) and the coordinates of all subsequent states can be found relative to this initial position based on the actions taken by the agent. We then create our state-transition graph out of these newly constructed features instead of the observations themselves. As the environment is discrete, we do not need to merge any of the states so we used a very small $\epsilon$=0.01 to create the graph. We created the graph out of 1,000,000 state-transitions from a random agent. The resulting graph had 42 different nodes. When the agent is in the starting position, using $n$=3 resulted in no risky states found which meant no direction of risk. Other values of n (up to $n$=10) also resulted in no risky states found. 

We then used LIME to find the direction of risk. We defined the one-hot perturbation vector as a vector of size 14 where the first 7 elements encoded a change in x by -3, -2, -1, 0, +1, +2, and +3. The other 7 elements did the same but for the y position. When creating the perturbations, we had one and only one element related to each feature be set to 1 at a time.
In other words, each dimension of the current location (x and y) is independently perturbed by adding a random integer number between [-3,3] to it.
If the perturbation resulted in a state inside a wall, we ignored that perturbation. If the perturbed state was in a lava tile, we used a label of 1 and otherwise a label of 0. The explanation offered by LIME in the same starting position was $g=(53.74, 0.0)$. This means a direction of risk corresponding to the increase of the x coordinate --i.e. pointing to the right. But due to the existence of the walls, such a change is actually impossible and our method's assessment that the agent's current situation is safe is a more logical one.

While this is definitely an extreme case, it shows why LIME or the other feature and perturbation based explanation methods are not suitable for RL environments. The information about state transitions is necessary for more accurate explanations in this space.

A final series of tests were performed on a real-world problem. RL methods have been proposed before for devising treatment strategies for ICU patients who match the sepsis criteria\cite{raghu2017continuous, saria2018individualized, raghu2018model}. As safety is very important in healthcare, we thought that this area could be a good candidate for a practical implementation of our method. We performed these tests on the MIMIC-III dataset\cite{mimiciii}. The data extraction and preparation and feature generation process was the same as \cite{raghu2017continuous}. To create the environment states for each patient, 47 health features were defined in 4-hour intervals during their stay in ICU. This data is then used to create the state-transition graph. Out of the approximately 270000 states recorded from the dataset, about 170000 graph nodes were created using $\epsilon=0.2$. In this experiment, we defined the risk function $R_i(s)$ as:

$R_{i+1}(s) = (1-l)*R_{i}(s)+l*\sqrt{\frac{\sum_{k\in nexts(s)}{R_{i}(k)^2*V(s,k)}}{\sum_{k\in nexts(s)}{V(s,k)}}},$

$R_0(s)=\left\{\begin{matrix}
0.5+\frac{s_{fatal}}{2*s_{total}} & s_{fatal} \neq 0\\ 
0 & otherwhise
\end{matrix}\right.$

\noindent where $s$ is a node in the state-transition graph, $l$ is a learning rate, $nexts(s)$ is the set of nodes in the graph that there is a directed edge going from $s$ to them, $V(s,k)$ is the number of directed edges that go from $s$ to $k$, $s_{fatal}$ is the number of states in the original dataset that the patient died in and also are represented in the graph by $s$, and $s_{total}$ is the number of the states in the original dataset that are represented by $s$.

The reasoning behind this risk function is that there are many patients in this environment that reached some states and died, while there are other patients that reached the same states but survived. This formulation turns the risk into a probabilistic function. It also enables nodes that did not observe fatal states but lead to other nodes that do, to be considered riskier than nodes that are far from any fatality, effectively letting the risk to grow the closer the states get to a fatal state. The reason for the risk of nodes with fatal states starting from 0.5 is to make sure that the risk signal does not get lost during the recursion process. We updated the risk function of each graph node using this formula for 50 iterations with $l=0.01$.

\begin{figure}[ht]
\begin{center}
\centerline{\includegraphics[width=\columnwidth]{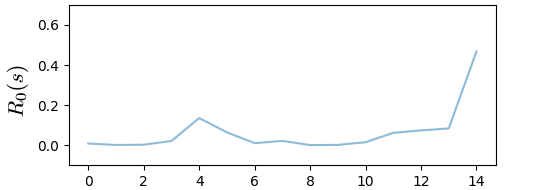}}
\centerline{\includegraphics[width=\columnwidth]{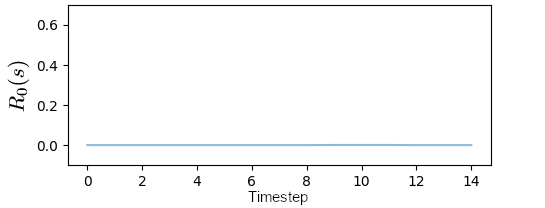}}
\caption{Risk over timesteps for two different patients in the MIMIC-III dataset. The vertical axis is the risk value $R_{50}(s)$ and the horizontal axis is timestep as defined during the data extraction process. The patient whose graph is on the top did not survive while the other one did. As can be seen, risk was evident even before the death of the patient.}
\label{f5}
\end{center}
\end{figure}

We then picked a number of patients and drew the risk graph per timestep during their stay in ICU. This process is not explaining the risks of an RL agent. Instead, it is trying to explain the risks of a human agent. While we saw differences between fatal and recovered cases as seen in Figure~\ref{f5}, we refrain from claiming success in this case because we have no robust way to prove the validity of it. Nevertheless, this experiment shows that our method is general enough that it is not limited to explaining risk of RL agents only. We also performed experiments with the direction of risk. To do so, we used linear regression instead of linear classification to find the direction of increasing risk. But to evaluate these explanations in a robust manner, real-world experiments need to be performed. Such experiments are not necessarily possible due to health and ethical concerns.

\section{Conclusion and Future Work}

In this paper, we proposed a method for finding the local direction of risk in reinforcement learning. This method is designed to work in environments with continuous state and action spaces so it does not fall into the common pitfalls of Explainable RL methods that cannot be used for many real-world situations which have continuous state and/or action spaces. It is also focusing on local explanations and objective functions like risk that are related to the use of the agent in the real world.

A future area of research can be in creating state transition graphs and models that are less resource-intensive. State-transition models that can even extrapolate to never-before-seen state regions are also a good avenue of further research. Neural networks trained to act as a state-transition model are a possible alternative which could potentially alleviate the discretization problem of the graph based ones. More work could also be done on offering versions of this method for other types of RL environments with multiple objective or risk functions or with multiple distinct risk areas.


\begin{thebibliography}{00}
\bibitem{molnar2020interpretable}
C.~Molnar, \emph{Interpretable Machine Learning}.\hskip 1em plus 0.5em minus
  0.4em\relax Lulu. com, 2020.

\bibitem{adadi2018peeking}
A.~Adadi and M.~Berrada, ``Peeking inside the black-box: A survey on
  explainable artificial intelligence (xai),'' \emph{IEEE Access}, vol.~6, pp.
  52\,138--52\,160, 2018.

\bibitem{ribeiro2016should}
M.~T. Ribeiro, S.~Singh, and C.~Guestrin, ``Why should i trust you?: Explaining
  the predictions of any classifier,'' in \emph{Proceedings of the 22nd ACM
  SIGKDD international conference on knowledge discovery and data
  mining}.\hskip 1em plus 0.5em minus 0.4em\relax ACM, 2016, pp. 1135--1144.

\bibitem{Bastani2018}
O.~Bastani, Y.~Pu, and A.~Solar-Lezama, ``{Verifiable reinforcement learning
  via policy extraction},'' \emph{Advances in Neural Information Processing
  Systems}, vol. 2018-Decem, no. NeurIPS, pp. 2494--2504, 2018.

\bibitem{rodriguez2019interpretable}
I.~D.~J. Rodriguez, T.~Killian, S.-H. Son, and M.~Gombolay, ``Interpretable
  reinforcement learning via differentiable decision trees,'' \emph{arXiv
  preprint arXiv:1903.09338}, 2019.

\bibitem{roth2019}

A.~M. Roth, N.~Topin, P.~Jamshidi, and M.~Veloso, ``Conservative q-improvement:
  Reinforcement learning for an interpretable decision-tree policy,''
  \emph{CoRR}, vol. abs/1907.01180, 2019. [Online]. Available:
  \url{http://arxiv.org/abs/1907.01180}


\bibitem{Maes2012}
F.~Maes, R.~Fonteneau, L.~Wehenkel, and D.~Ernst, ``{Policy search in a space
  of simple closed-form formulas: Towards interpretability of reinforcement
  learning},'' \emph{Lecture Notes in Computer Science (including subseries
  Lecture Notes in Artificial Intelligence and Lecture Notes in
  Bioinformatics)}, vol. 7569 LNAI, pp. 37--51, 2012.

\bibitem{hein2018interpretable}
D.~Hein, S.~Udluft, and T.~A. Runkler, ``Interpretable policies for
  reinforcement learning by genetic programming,'' \emph{Engineering
  Applications of Artificial Intelligence}, vol.~76, pp. 158--169, 2018.

\bibitem{Topin2019}
N.~Topin and M.~Veloso, ``{Generation of Policy-Level Explanations for
  Reinforcement Learning},'' \emph{Proceedings of the AAAI Conference on
  Artificial Intelligence}, vol.~33, no.~2, pp. 2514--2521, 2019.

\bibitem{samsudin2011highly}
K.~Samsudin, F.~A. Ahmad, and S.~Mashohor, ``A highly interpretable fuzzy rule
  base using ordinal structure for obstacle avoidance of mobile robot,''
  \emph{Applied Soft Computing}, vol.~11, no.~2, pp. 1631--1637, 2011.

\bibitem{Hein2017}
D.~Hein, A.~Hentschel, T.~Runkler, and S.~Udluft, ``{Particle swarm
  optimization for generating interpretable fuzzy reinforcement learning
  policies},'' \emph{Engineering Applications of Artificial Intelligence},
  vol.~65, no. July, pp. 87--98, 2017. [Online]. Available:
  \url{http://dx.doi.org/10.1016/j.engappai.2017.07.005}


\bibitem{Beyret2019}
B.~Beyret, A.~Shafti, and A.~A. Faisal, ``{Dot-to-Dot: Explainable Hierarchical
  Reinforcement Learning for Robotic Manipulation},'' 2019. [Online].
  Available: \url{http://arxiv.org/abs/1904.06703}


\bibitem{Lyu2019}
D.~Lyu, F.~Yang, B.~Liu, and S.~Gustafson, ``{SDRL: Interpretable and
  Data-Efficient Deep Reinforcement Learning Leveraging Symbolic Planning},''
  \emph{Proceedings of the AAAI Conference on Artificial Intelligence},
  vol.~33, pp. 2970--2977, 2019.

\bibitem{Liu2019}
G.~Liu, O.~Schulte, W.~Zhu, and Q.~Li, ``{Toward interpretable deep
  reinforcement learning with linear model u-trees},'' \emph{Lecture Notes in
  Computer Science (including subseries Lecture Notes in Artificial
  Intelligence and Lecture Notes in Bioinformatics)}, vol. 11052 LNAI, pp.
  414--429, 2019.

\bibitem{Verma2018}
A.~Verma, V.~Murali, R.~Singh, P.~Kohli, and S.~Chaudhuri, ``{Programmatically
  interpretable reinforcement learning},'' \emph{35th International Conference
  on Machine Learning, ICML 2018}, vol.~11, pp. 8024--8033, 2018.

\bibitem{Amir2018}
O.~Amir, F.~Doshi-Velez, and D.~Sarne, ``{Agent Strategy Summarization},''
  \emph{Proceedings of the 17Th International Conference on Autonomous Agents
  and Multiagent Systems (Aamas' 18)}, pp. 1203--1207, 2018.

\bibitem{Lage2019}
I.~Lage, D.~Lifschitz, F.~Doshi-Velez, and O.~Amir, ``{Exploring Computational
  User Models for Agent Policy Summarization},'' no.~3, pp. 1401--1407, 2019.

\bibitem{juozapaitis2019explainable}
Z.~Juozapaitis, A.~Koul, A.~Fern, M.~Erwig, and F.~Doshi-Velez, ``Explainable
  reinforcement learning via reward decomposition,'' in \emph{Proceedings of
  the IJCAI 2019 Workshop on Explainable Artificial Intelligence}, 2019, pp.
  47--53.

\bibitem{selvaraju2017grad}
R.~R. Selvaraju, M.~Cogswell, A.~Das, R.~Vedantam, D.~Parikh, and D.~Batra,
  ``Grad-cam: Visual explanations from deep networks via gradient-based
  localization,'' in \emph{Proceedings of the IEEE International Conference on
  Computer Vision}, 2017, pp. 618--626.

\bibitem{Weitkamp2019}
L.~Weitkamp, E.~van~der Pol, and Z.~Akata, ``{Visual Rationalizations in Deep
  Reinforcement Learning for Atari Games},'' pp. 151--165, 2019.

\bibitem{VanderWaa2018}
J.~van~der Waa, J.~van Diggelen, K.~van~den Bosch, and M.~Neerincx,
  ``{Contrastive Explanations for Reinforcement Learning in terms of Expected
  Consequences},'' 2018. [Online]. Available:
  \url{http://arxiv.org/abs/1807.08706}

\bibitem{Pecka2014}
M.~Pecka and T.~Svoboda, ``{Safe exploration techniques for reinforcement
  learning – an overview},'' \emph{Lecture Notes in Computer Science
  (including subseries Lecture Notes in Artificial Intelligence and Lecture
  Notes in Bioinformatics)}, vol. 8906, pp. 357--375, 2014.

\bibitem{metzen2013learning}
J.~H. Metzen, ``Learning graph-based representations for continuous
  reinforcement learning domains,'' in \emph{Joint European Conference on
  Machine Learning and Knowledge Discovery in Databases}.\hskip 1em plus 0.5em
  minus 0.4em\relax Springer, 2013, pp. 81--96.

\bibitem{8575599}
M.~{Ribeiro Furtado de Mendonça}, A.~{Ziviani}, and A.~{da Motta Salles
  Barreto}, ``Abstract state transition graphs for model-based reinforcement
  learning,'' in \emph{2018 7th Brazilian Conference on Intelligent Systems
  (BRACIS)}, 2018, pp. 115--120.

\bibitem{brockman2016openai}
G.~Brockman, V.~Cheung, L.~Pettersson, J.~Schneider, J.~Schulman, J.~Tang, and
  W.~Zaremba, ``Openai gym,'' \emph{arXiv preprint arXiv:1606.01540}, 2016.

\bibitem{fujimoto2018addressing}
S.~Fujimoto, H.~van Hoof, and D.~Meger, ``Addressing function approximation
  error in actor-critic methods,'' \emph{arXiv preprint arXiv:1802.09477},
  2018.

\bibitem{gym_minigrid}
M.~Chevalier-Boisvert, L.~Willems, and S.~Pal, ``Minimalistic gridworld
  environment for openai gym,'' \url{https://github.com/maximecb/gym-minigrid},
  2018.

\bibitem{raghu2017continuous}
A.~Raghu, M.~Komorowski, L.~A. Celi, P.~Szolovits, and M.~Ghassemi,
  ``Continuous state-space models for optimal sepsis treatment-a deep
  reinforcement learning approach,'' \emph{arXiv preprint arXiv:1705.08422},
  2017.

\bibitem{saria2018individualized}
S.~Saria, ``Individualized sepsis treatment using reinforcement learning,''
  \emph{Nature medicine}, vol.~24, no.~11, pp. 1641--1642, 2018.

\bibitem{raghu2018model}
A.~Raghu, M.~Komorowski, and S.~Singh, ``Model-based reinforcement learning for
  sepsis treatment,'' \emph{arXiv preprint arXiv:1811.09602}, 2018.

\bibitem{mimiciii}
A.~E. Johnson, T.~J. Pollard, L.~Shen, H.~L. Li-wei, M.~Feng, M.~Ghassemi,
  B.~Moody, P.~Szolovits, L.~A. Celi, and R.~G. Mark, ``Mimic-iii, a freely
  accessible critical care database,'' \emph{Scientific data}, vol.~3, p.
  160035, 2016.

\end{thebibliography}
\end{document}